\documentclass{bmvc2k}

\title{Spatio-Temporal and Clinical Conditioning for Fine-Grained Radiology Report Retrieval}

\addauthor{Phillip Sloan}{phillip.sloan@bristol.ac.uk}{1}
\addauthor{Edwin Simpson}{edwin.simpson@bristol.ac.uk}{1}
\addauthor{Majid Mirmehdi}{m.mirmehdi@bristol.ac.uk}{1}

\addinstitution{
 Department of Computer Science\\
 University of Bristol\\
 Bristol, UK
}

\runninghead{Student, Prof, Collaborator}{BMVC Author Guidelines}

\usepackage{amsmath}
\usepackage{amsfonts}
\usepackage{graphicx}
\usepackage{needspace}
\usepackage{makecell}
\usepackage[compatibility=false]{caption}

\definecolor{bmv@captioncolor}{rgb}{0,0,.4}

\usepackage{xspace}
\usepackage[table]{xcolor}

\newcommand{\framework}{STAR3\xspace}

\begin{document}
\maketitle

\begin{abstract}
Radiology is vital to modern healthcare, but rising imaging demand and persistent workforce shortages strain reporting capacity and clinical workflows. Automated radiology report generation has the potential to support radiologists and help alleviate this burden; however, existing retrieval-based methods remain rigid, lack explicit anatomical grounding, and do not account for longitudinal disease progression or available clinical context. In this work, we introduce \framework, a multimodal, spatio-temporal, attentive retrieval framework for radiology report generation that aligns region-level anatomical information with clinical indications and longitudinal changes across chest X-ray studies. Our framework employs an object detector to identify anatomically meaningful regions and retrieves semantically relevant report sentences conditioned on both current clinical context and changes observed between prior and current examinations. This design enables anatomically and temporally grounded report generation that better reflects clinical reporting practice. Experiments on the MIMIC-CXR dataset demonstrate that \framework outperforms current retrieval-based approaches on retrieval, NLP and clinical metrics, highlighting the value of conditioning retrieval anatomically, temporally and clinically for advancing automated radiology report generation.
\end{abstract}
\section{Introduction}
Radiology is essential to modern healthcare, enabling clinicians to diagnose disease and monitor treatment outcomes. Radiology includes various imaging modalities, however the most common examination performed is the chest x-ray which is critical for identifying common thoracic diseases such as pneumonia and lung cancer \cite{MIMIC}. At the same time, the growing disparity between the capacity of the radiology workforce and service demand has become a major concern, with nearly all (99\%) clinical directors expressing concern about the impact on the safety and quality of care \cite{RCR}. To address this challenge, AI-based approaches have been increasingly identified as a promising means to augment radiologists’ reporting workflows \cite{RCR2}, motivating a rapid research interest in automated radiology report generation (ARRG) as a means to support clinical practice \cite{Sloan2024}.

\noindent
\begin{minipage}[t]{0.44\textwidth}
Most ARRG research has focused on autoregressive methodologies \cite{tanida2023, Sloan2024,bannur2024maira2,Liu_2025_CVPR}, which, despite producing fluent reports, often hallucinate clinically incorrect findings \cite{Yan2024} while also becoming computationally intensive due to the adoption of large language models (LLMs) into the architecture. Retrieval-based approaches have gained traction as an alternative methodology as they constrain outputs to a corpus of human-written reports, reducing the chance of hallucinated content \cite{endo21,Yan2024} while achieving competitive performance without dependence on large-parameter generative models \cite{endo21}.
\end{minipage}
\hfill
\begin{minipage}[t]{0.52\textwidth}
\vspace{-4pt}
\centering
\includegraphics[width=\linewidth]{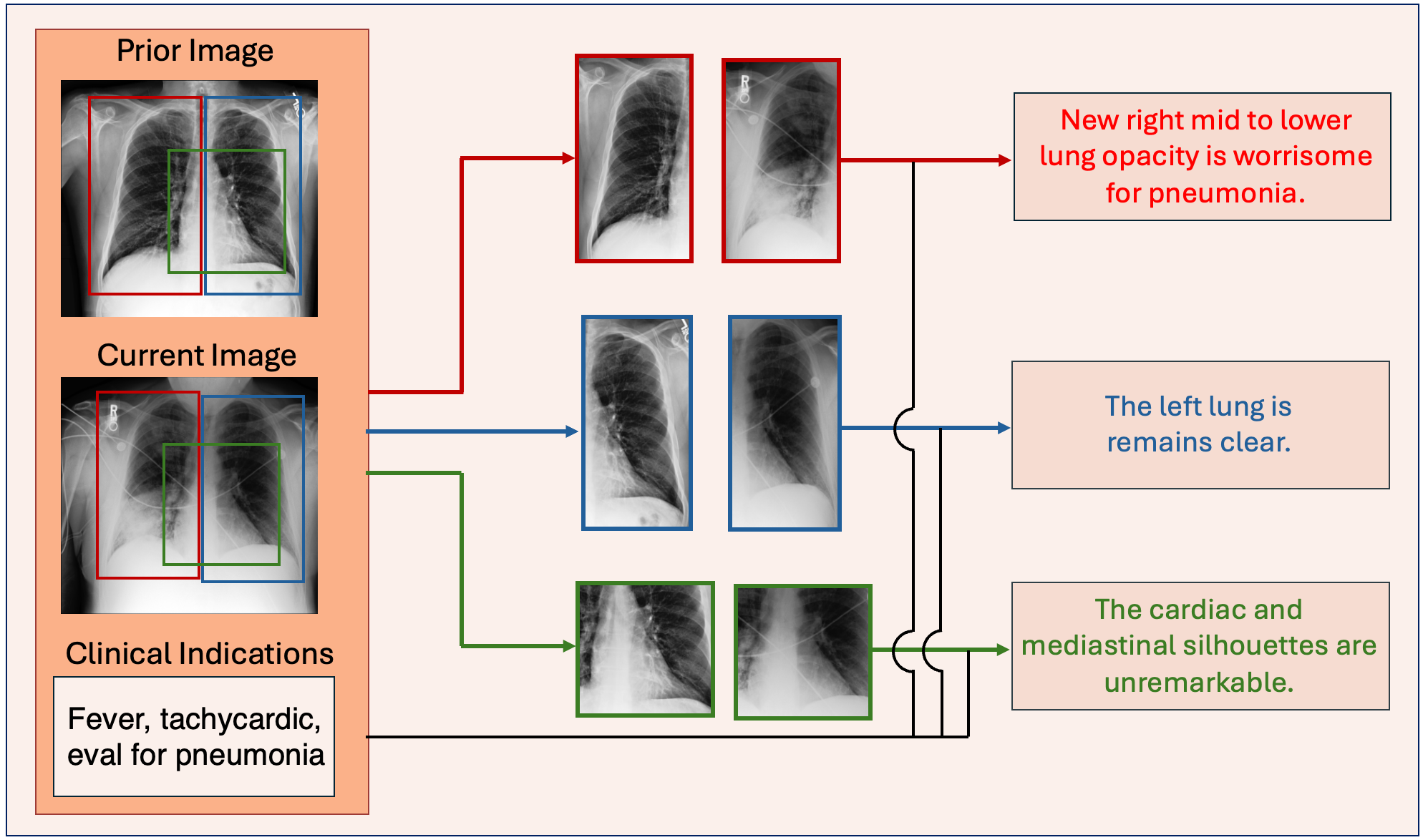}
\captionsetup{
  type=figure,
  skip=-1pt
}
\captionof{figure}{Conceptual overview of STAR3. The framework compares current and prior chest X-rays at the anatomical-region level, incorporates the clinical indication, and retrieves region-specific report sentences that capture anatomically localised findings and temporal change.}
\label{fig:conceptual_overview}
\end{minipage}

\medskip

Current retrieval-based approaches exhibit several limitations. Most methods retrieve a radiology report from a fixed corpus by selecting the instance with the highest image–text similarity to a queried image \cite{Boag2020,wang-etal-2022-medclip,Jeong2023MIDL,Zhang2025RadIR}. This strategy is inherently rigid as it restricts the retrieved output to existing reports and limits the model’s ability to accommodate subtle variations in report structure or content. Moreover, the temporal nature of radiology studies is often ignored with some works proposing to remove references to prior studies altogether \cite{Ramesh2022}, thereby discarding clinically meaningful longitudinal information. Such information is crucial for monitoring disease progression or improvement over time. Another limitation in retrieval methods is the under use of complementary clinical information by most existing approaches \cite{vanSonsbeek2023,Jeong2023MIDL,Zhang2025RadIR}. Clinical indications are readily available at the time of reporting and have been shown to enhance diagnostic accuracy in radiological practice \cite{Castillo2021}.

To address these limitations, we propose \framework (Spatio-Temporal Attention-based Radiology Report Retrieval), a novel multimodal, anatomically and temporally grounded retrieval framework for radiology report generation. Inspired by the clinical workflow of radiologists, who interpret findings at the level of anatomical regions in the context of patient history and longitudinal change \cite{tanida2023, sloan2025camchex}, \framework leverages region-level features extracted via an object detector \cite{FasterRCNN, raddino} trained on the Chest ImaGenome dataset \cite{chest-imagenome-1.0.0}. For each anatomical region, the model uses a learned gating mechanism to selectively enable sentence retrieval, retrieving candidate sentences from a shared embedding space structured by semi-supervised multi-modal contrastive learning and conditioned on both clinical indications and region-level temporal changes derived from paired current and prior chest X-rays. This design enables the retrieval of findings that are explicitly grounded in both anatomy and disease progression (e.g., distinguishing new, resolving, or stable abnormalities within specific regions).

Our contributions can be summarised as follows: (i) We propose \framework, a novel multimodal retrieval framework for radiology report generation that provides fine-grained anatomical grounding via region-level features extracted using an object detector trained on {frontal chest x-rays}.
(ii) We present, to our knowledge, the first retrieval-based report generation pipeline to perform temporally grounded sentence retrieval at the anatomical (region) level by jointly modelling current and prior chest X-rays, enabling explicit reasoning over longitudinal changes in localised findings. (iii) We incorporate an anatomical dropout module that selectively filters region-level predictions to reduce redundancy and enforce anatomical consistency in retrieved sentences. (iv) We introduce a semi-supervised contrastive learning objective that aligns image regions and report sentences using fine-grained clinical labels seeking to structure the retrieval space around clinically meaningful concepts and state changes. (v) We conduct extensive experiments on the MIMIC-CXR dataset, demonstrating that \framework consistently outperforms existing retrieval-based baselines, with ablation studies validating the contributions of anatomical grounding, temporal modelling, anatomical dropout, and semi-supervised contrastive learning.

\section{Related Works}
Our work is most closely related to retrieval-based ARRG methods that align visual representations of chest x-rays with report-level textual embeddings. We review these approaches together with prior work on temporal reasoning in longitudinal chest imaging, since \framework combines anatomically grounded region-level retrieval with explicit longitudinal modelling.

\subsection{Contrastive Learning for Radiology Report Retrieval}
Within ARRG, contrastive learning has been adopted for two main objectives: improving multimodal conditioning within encoder-decoder generation frameworks \cite{Hou2023, Li_2023_ICCV}, and enabling retrieval-based report generation through discriminative image-text embedding spaces. Retrieval-based methods typically generate reports by selecting semantically similar reports or textual units from a fixed radiological corpus. Boag et al. \cite{Boag2020} proposed one of the earliest retrieval approaches in ARRG, obtaining the report associated with the training image with the highest cosine similarity to a given test image. Subsequent work adopted CLIP-style vision-language approaches \cite{Radford2021CLIP}, replacing image-to-image retrieval with image-to-text retrieval \cite{wang-etal-2022-medclip,endo21,CLIP-XRad, UrRahman2025}. Van Sonsbeek and Worring~\cite{vanSonsbeek2023} introduced X-TRA, which used CLIP-based cross-modal retrieval augmentation to incorporate similar chest X-rays and reports through multi-head attention for disease classification and report retrieval. Jeong et al. \cite{Jeong2023MIDL} proposed X-REM, replacing cosine similarity with a learned multimodal matching score from a vision-language model.

More recent approaches have extended traditional image-to-text retrieval by incorporating disease-aware semantics, topic structure, or anatomical information. Zhou et al. \cite{Zhou2025} proposed HOPAN, which modeled hierarchical region-organ-global interactions between visual features and report tokens. Zhang et al. \cite{Zhang2025RadIR} introduced RadIR, which leveraged anatomy-conditioned similarity derived from radiology reports to support fine-grained retrieval. Zhao et al.~\cite{Teaser} introduced Teaser, a topic-aware retrieval framework that separates common and rare clinical concepts and uses topic contrastive learning to improve alignment with low-frequency but clinically important findings. Yan et al. \cite{Yan2024} developed AHIVE, which provided hierarchical, anatomy-aware visual embeddings that incorporate diagnostic structure into image-report retrieval. 

While these methods improve retrieval by incorporating clinical, topic-level, or anatomical structure, they primarily operate using global image-report representations and do not perform temporally grounded sentence retrieval at the anatomical region level.

\subsection{Temporal Reasoning} 
Temporal information is fundamental to medical image interpretation, providing important context for reasoning about disease progression. However, most existing retrieval-based report generation methods do not explicitly model the temporal structure of radiology reporting, often retrieving sentences that reference prior examinations even when no such prior is available. Ramesh et al. \cite{Ramesh2022} sought to tackle this issue by demonstrating that removing prior references from MIMIC-CXR \cite{MIMIC} improves retrieval performance, however, such preprocessing discards temporal information altogether, precluding longitudinal reasoning. In contrast, Bannur et al. \cite{Bannur2023} extended BioViL \cite{Boecking2022} with temporal modelling by including prior imaging. Our work builds upon this temporal perspective, but instead of retrieving a radiology report, we perform anatomically grounded, region-level sentence retrieval. This finer-grained design enables more precise and contextually appropriate retrieval, while label-supervised contrastive learning over Chest ImaGenome annotations further structures the retrieval space around clinically meaningful and temporally discriminative concepts.

Overall, existing retrieval-based methods do not jointly model anatomical grounding, region-level temporal change, and selective sentence retrieval within a unified framework. Our work is the first to combine these components in a retrieval-based report generation pipeline, enabling anatomically precise and temporally grounded sentence retrieval conditioned on clinical context. 

\section{Methodology}


We present \framework, a multimodal retrieval framework for automatic radiology report generation that integrates longitudinal, region-level visual characteristics with textual clinical indications. Moving away from traditional global image-report matching, our approach isolates specific anatomical regions, tracking their evolution across time and filtering out non-essential anatomical contexts before executing cross-modal retrieval.

\begin{figure*}[t]
\centering
\includegraphics[width=.92\textwidth]{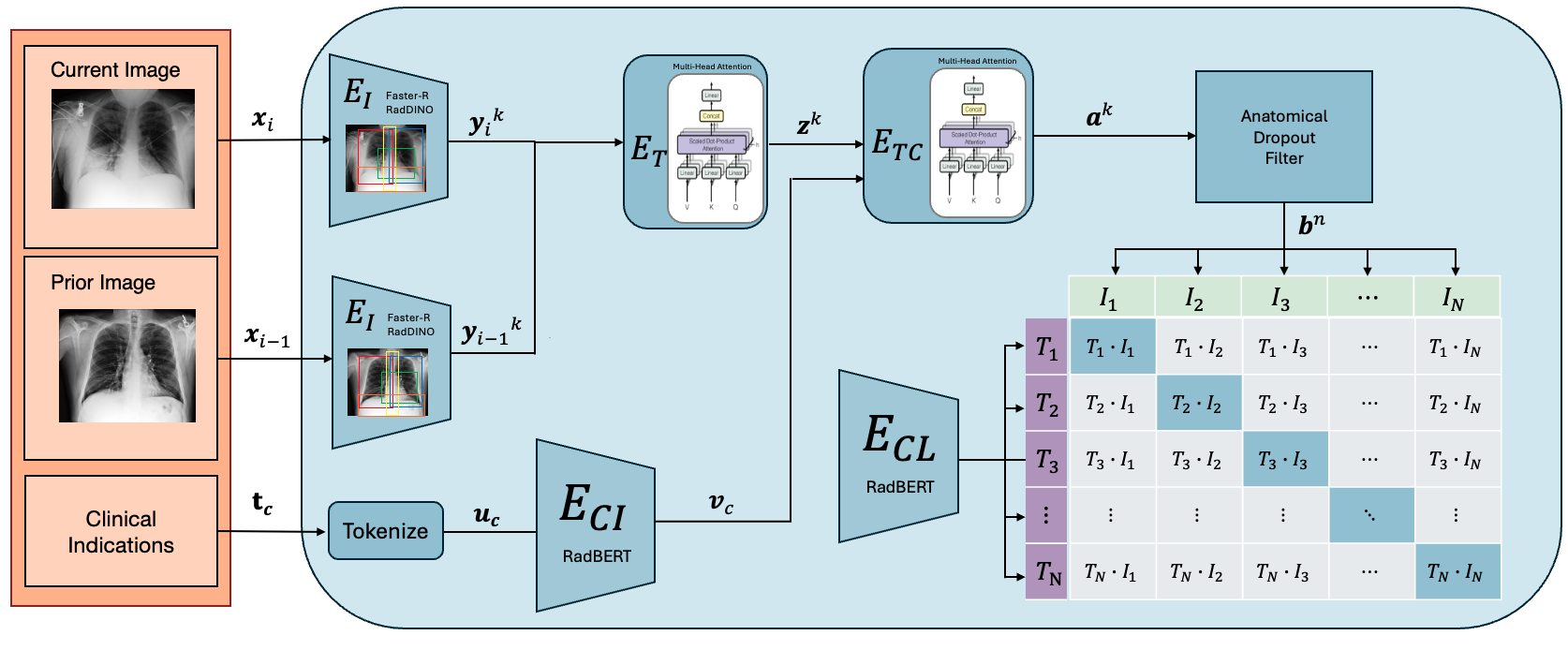}
\vspace{2mm}
\caption{Overview of the proposed \framework architecture. The framework jointly processes current and prior chest X-ray images together with associated clinical indications to learn temporally grounded and clinically conditioned region-level representations. Anatomical features extracted using a RadDINO-based detector are fused through transformer-based temporal and clinical cross-attention modules to model longitudinal anatomical changes and clinical context. An auxiliary abnormality prediction branch and anatomical dropout module are incorporated to identify clinically relevant regions prior to region-level contrastive retrieval between anatomical descriptions and localised image representations.}
\label{fig:architecture}
\end{figure*}

As illustrated in Figure~\ref{fig:architecture}, the framework follows a modular design comprising four core elements: (i) anatomical region detection and domain-specific feature extraction, (ii) region-level temporal modelling, (iii) clinical conditioning via cross-modal attention, followed by a multi-task \textit{anatomical dropout} module, and (iv) semi-supervised cross-modal contrastive learning to structure the joint retrieval embedding space.

\subsection{Anatomical Region Detection and Feature Extraction} \label{sec:anatomicalregiondetection}

Given a chest X-ray image $x \in \mathbb{R}^{C \times H \times W}$, we use a Faster R-CNN-style detector \cite{FasterRCNN} to identify $K$ predefined anatomical regions and extract region-level visual features. Rather than using the conventional ResNet-50 backbone \cite{ResNet}, we employ RadDINO \cite{raddino}, a biomedical vision encoder pretrained solely on unimodal biomedical imaging data, as the image backbone. The input image $x$ is divided into non-overlapping patches of size $P \times P$, producing $M = (H/P)(W/P)$ image patches. RadDINO maps these patches to a sequence of patch embeddings $\mathbf{T} = f_{\text{RadDINO}}(x) \in \mathbb{R}^{M \times D_{\text{rad}}}$, where $D_{\text{rad}}$ is the RadDINO embedding dimensionality. Since the patch embeddings preserve the spatial ordering of the original image, we reshape $\mathbf{T}$ into a two-dimensional feature map $\mathbf{F} \in \mathbb{R}^{D_{\text{rad}} \times H' \times W'}$, where $H' = H/P$ and $W' = W/P$. This reconstructed spatial feature map is then used by the detection head to generate anatomical region proposals, each associated with a bounding box and class score. Since RadDINO and the Faster R-CNN detection heads operate with different feature dimensions, we use a $1\times1$ convolution to map the RadDINO patch feature map to a detector-compatible feature map $\mathbf{F}_{\text{det}} \in \mathbb{R}^{{D_{\text{det}}} \times H' \times W'}$, where $D_{\text{det}}$ is the detector feature dimensionality. 

Following the Faster R-CNN framework~\cite{FasterRCNN}, the RPN and RoI heads operate on $\mathbf{F}_{\text{det}}$ to produce candidate bounding boxes and anatomical class scores. Let $\{(\hat{\mathbf{b}}_i,\hat{\mathbf{p}}_i)\}_{i=1}^{N_{\text{cand}}}$ denote the set of candidate detections, where $\hat{\mathbf{b}}_i=(x_i^1, y_i^1, x_i^2, y_i^2)$ is the predicted bounding box and $\hat{\mathbf{p}}_i \in [0,1]^K$ contains the confidence scores for the $K$ predefined anatomical classes. For each anatomical class $k \in \{1,\ldots,K\}$, we select the highest-confidence proposal, $\boldsymbol{\beta}_k=\hat{\mathbf{b}}_{i^\ast}$ where $i^\ast=\arg\max_i \hat{p}_{i,k}$, and discard classes with no valid detection. The corresponding RoI feature map is extracted from $\mathbf{F}_{\text{det}}$, spatially pooled, and projected to obtain the region-level representation $\mathbf{y}_k\in\mathbb{R}^{D_{\text{roi}}}$, where $D_{\text{roi}}$ denotes the dimensionality of the projected RoI features. The detector therefore outputs a set of detected anatomical regions, each represented by a bounding box $\boldsymbol{\beta}_k$ and corresponding feature vector $\mathbf{y}_k$. Across all detected regions, these feature vectors can be written as $\mathbf{Y}\in\mathbb{R}^{N_{\text{det}}\times D_{\text{roi}}}$, where $N_{\text{det}}\leq K$.

\subsection{Region-Level Temporal Modelling} \label{sec:reg-lv-tmp-modelling}

By processing both current and prior studies, the model can reason over longitudinal anatomical and pathological changes. To capture these changes, \framework fuses current and prior chest X-rays at the anatomical level, producing temporally enriched anatomical features.

Formally, we apply our object detector $\mathbf{E}_I$ from Section~\ref{sec:anatomicalregiondetection} to both the current and prior chest X-rays, ensuring that region-level visual representations are extracted over a consistent set of anatomical regions across examinations. Let $\mathbf{y}_k^{t-1}$ and $\mathbf{y}_k^{t}$ denote the region-level embeddings for anatomical region $k$ from the prior and current examinations, respectively, where $\mathbf{y}_k^{t-1}, \mathbf{y}_k^{t} \in \mathbb{R}^{D_{\text{roi}}}$. If a prior study is unavailable, $\mathbf{y}_k^{t-1}$ is replaced with a learnable embedding. 

We model temporal change at the anatomical level using a multi-head attention (MHA) module. For each anatomical region $k$, the current representation $\mathbf{y}_k^{t}$ attends to both the current and prior region embeddings, $\mathbf{y}_k^{t}$ and $\mathbf{y}_k^{t-1}$, enabling the model to integrate prior anatomical context while preserving focus on the current examination. Specifically, $\mathbf{y}_k^{t}$ is used as the query, while $\{\mathbf{y}_k^{t}, \mathbf{y}_k^{t-1}\}$ are used as the keys and values. This produces a temporally enriched region representation, $\mathbf{z}_k = f_{\text{temporal}}(\mathbf{y}_k^{t}, \mathbf{y}_k^{t-1})$,
which encodes region-specific longitudinal information anchored to the current examination. These temporally enriched region features are used as inputs to the subsequent clinical conditioning module.

\subsection{Clinical Conditioning}

Although the temporally enriched anatomical features $\mathbf{z}_k$ encode region-specific longitudinal information, they do not explicitly capture the clinical context motivating the examination. To incorporate this diagnostic context, we utilise clinical indications, which in practice are provided to radiologists prior to image interpretation and describe the symptoms, suspected conditions, or clinical questions motivating the examination.

Formally, we tokenise the clinical indications $\mathbf{t}_c$ into $\mathbf{u}_c$ and encode them using RadBERT \cite{Radford2021CLIP}, a clinical language transformer pretrained on radiology reports, yielding contextual text embeddings $\mathbf{V}_c$. To incorporate this clinical context, we apply region-to-text multi-head cross-attention between the temporally enriched anatomical representations and the indication embeddings. For each anatomical region $k$, the temporal representation $\mathbf{z}_k$ attends over the clinical indication embeddings $\mathbf{V}_c$, allowing the model to emphasise textual context relevant to that region. This produces a clinically conditioned anatomical representation $\mathbf{a}_k = f_{\text{clinical}}(\mathbf{z}_k, \mathbf{V}_c)$, which combines region-specific longitudinal visual information with the clinical context motivating the examination. The function $f_{\text{clinical}}$ is implemented using multi-head cross-attention in the same fashion as in Section \ref{sec:reg-lv-tmp-modelling}.

\subsection{Anatomical Dropout Filter}

While the clinically conditioned anatomical representations $\mathbf{a}_k$ encode anatomical, temporal, and indication-specific information, not every detected region is relevant to the final report. Radiology reports are inherently selective, typically describing only regions that contain clinically meaningful findings or provide necessary context. Naively retrieving sentences for every detected anatomical region can therefore introduce redundant, irrelevant, or anatomically inconsistent content. To address this, we introduce an anatomical dropout module that acts as a region-level content selection mechanism, retaining only anatomical regions predicted to contribute to the report.

Formally, given the clinically conditioned anatomical representation $\mathbf{a}_k$ for each detected region $k$, the anatomical dropout module predicts a binary region-selection label $s_k \in \{0,1\}$ indicating whether that region should contribute a sentence to the final report. In parallel, following Tanida et al.~\cite{tanida2023}, we train an auxiliary pathology classifier to predict an abnormality label $o_k \in \{0,1\}$, indicating whether region $k$ contains abnormal findings. The selection and pathology heads are implemented as lightweight multi-layer perceptrons and are trained jointly with the preceding temporal fusion and clinical conditioning modules.

At inference time, the region-selection probabilities are first used to filter the full set of anatomical representations $\{\mathbf{a}_k\}_{k=1}^{K}$ into a smaller set of report-relevant representations $\{\mathbf{b}_n\}_{n=1}^{N_{\text{sel}}}$, where $N_{\text{sel}} \leq K$. The pathology predictions are then used to constrain sentence retrieval for each retained region: regions predicted as normal retrieve only from the normal sentence embedding space, while regions predicted as abnormal retrieve only from the abnormal sentence embedding space. This pathology-aware filtering reduces redundant and anatomically inconsistent retrieval while encouraging each selected region to retrieve sentences with the appropriate clinical status.

\subsection{Self-Supervised Multi-Modal Contrastive Learning}

To align anatomical region representations with their corresponding report sentences, we introduce a self-supervised multi-modal contrastive learning objective that structures a shared embedding space between region-level visual features and region-level textual descriptions. This objective complements the temporal fusion, clinical conditioning, and anatomical dropout modules by encouraging anatomically matched region-sentence pairs to be close in embedding space, while mismatched pairs are pushed apart.

Given the selected anatomical representations $\{\mathbf{b}_n\}_{n=1}^{N_{\text{sel}}}$, each region representation is projected into a shared embedding space using a learnable visual projection head. Each report region description is independently encoded using RadBERT and projected into the same space using a learnable text projection head. The resulting visual and textual embeddings are $\ell_2$-normalised prior to similarity computation. We train this shared embedding space using four complementary objectives. First, we use a global bidirectional contrastive loss, $\mathcal{L}_{\text{global}}$, where each projected region embedding is encouraged to be most similar to its paired sentence embedding relative to other sentence embeddings in the mini-batch, and vice versa.

Formally, the corresponding similarity matrix is computed as $s_{ij}=\mathbf{c}_i^{\top}\mathbf{e}_j/\tau$, where $\tau$ is the contrastive temperature, and the global loss is optimised using the bidirectional cross-entropy objective,
\begin{equation}
\mathcal{L}_{\text{global}}=
\frac{1}{2N_{\text{pair}}}\sum_{i=1}^{N_{\text{pair}}}
\left[
-\log\frac{\exp(s_{ii})}{\sum_{j=1}^{N_{\text{pair}}}\exp(s_{ij})}
-\log\frac{\exp(s_{ii})}{\sum_{j=1}^{N_{\text{pair}}}\exp(s_{ji})}
\right],
\label{eq:contrastiveloss}
\end{equation}
where $N_{\text{pair}}$ is the number of matched region-sentence pairs in the mini-batch, $s_{ii}$ denotes the similarity of the matched pair, and the off-diagonal similarities $s_{ij}$ and $s_{ji}$ act as in-batch negatives for the region-to-text and text-to-region directions, respectively.

Second, we introduce a local token-level contrastive loss, $\mathcal{L}_{\text{local}}$, to encourage finer-grained alignment between anatomical regions and clinically relevant words or phrases. Rather than comparing each region only with a global sentence embedding, we compute similarities between each region embedding and the token-level embeddings of each candidate sentence. These token similarities are aggregated using a smooth maximum, allowing the model to attend to the most relevant textual tokens for each anatomical region. The resulting token-aware similarity is defined as:
\begin{equation}
s^{\text{local}}_{ij}
=
\tau_{\text{local}}\log\sum_{m=1}^{L_j}
\exp\left(
\frac{\mathbf{c}_i^{\top}\mathbf{d}_{j,m}}{\tau_{\text{local}}}
\right),
\end{equation}
where $\mathbf{d}_{j,m}$ denotes the $m$-th token embedding of candidate sentence $j$, $L_j$ is the number of tokens, and $\tau_{\text{local}}$ controls the smooth maximum used to aggregate token-level similarities. The local contrastive loss $\mathcal{L}_{\text{local}}$ follows the same bidirectional cross-entropy objective as Equation~\ref{eq:contrastiveloss}, replacing the global similarity $s_{ij}$ with the token-aware similarity $s^{\text{local}}_{ij}$.

Third, similarly to Jeong et al.~\cite{Jeong2023MIDL} and Smit et al.~\cite{smit-etal-2020-combining}, we utilise hard negatives loss {$\mathcal{L}_{\text{hn}}$} to improve retrieval specificity. For each anatomical region, we define hard negatives as anatomical region descriptions with highly overlapping structured Chest ImaGenome \cite{chest-imagenome-1.0.0} labels, but with the polarity of at least one finding reversed. Each signature consists of finding-relation pairs indicating whether a finding is present or absent. Given an anchor pair, we identify candidate negatives from the same anatomical region by flipping one finding relation while preserving as much of the remaining clinical context as possible. These hard negatives are therefore semantically close to the anchor but clinically incorrect. We penalise cases where such negatives are scored too close to, or higher than, the matched sentence, encouraging the model to learn fine-grained distinctions between similar region-level descriptions. The hard-negative loss is implemented as the margin-based objective
\begin{equation}
\mathcal{L}_{\text{hn}} = \frac{1}{N_{\text{hn}}}\sum_{q=1}^{N_{\text{hn}}}\max\left(0, \gamma - (s_q^{+}-s_q^{-})\right),
\end{equation}
where $N_{\text{hn}}$ is the number of mined hard-negative pairs, $s_q^{+}$ is the similarity score of the matched pair, $s_q^{-}$ is the similarity score of the corresponding hard-negative pair, and $\gamma$ is the margin.

Finally, we add an image-text matching (ITM) loss, $\mathcal{L}_{\text{ITM}}$, that directly predicts whether a candidate region-sentence pair is matched. The ITM head receives the concatenation of the projected region and sentence embeddings and is trained with a binary classification objective over positive matched pairs and in-batch negative pairs. This provides an additional pairwise alignment signal beyond contrastive similarity. The ITM loss is defined as a binary cross-entropy objective,
\begin{equation}
\mathcal{L}_{\text{ITM}} = -\frac{1}{N_{\text{ITM}}}
\sum_{r=1}^{N_{\text{ITM}}}
\left[ g_r \log \sigma(a_r) + (1-g_r)\log(1-\sigma(a_r)) \right],
\end{equation}
where $a_r=f_{\text{ITM}}([\mathbf{c}_r;\mathbf{e}_r])$ is the ITM logit for a candidate region-sentence pair, $g_r$ indicates whether the pair is matched, and $N_{\text{ITM}}$ is the number of ITM training pairs.

The global and local contrastive losses provide complementary alignment signals: $\mathcal{L}_{\text{global}}$ aligns projected region embeddings with sentence-level embeddings, while $\mathcal{L}_{\text{local}}$ encourages finer-grained alignment with clinically relevant token-level embeddings. The hard-negative and ITM losses provide additional supervision from clinically similar mismatches and explicit region-sentence matching, respectively.

To make the embedding space pathology-aware, we use abnormality labels to constrain valid region-sentence comparisons during contrastive learning. Region-sentence pairs with inconsistent abnormality status are masked, encouraging normal regions to align with normal descriptions and abnormal regions to align with abnormal descriptions. The resulting alignment objective is defined as

\begin{equation}
\mathcal{L}_{\text{align}} =
\mathcal{L}_{\text{global}} +
\mathcal{L}_{\text{local}} +
\mathcal{L}_{\text{hn}} +
\mathcal{L}_{\text{ITM}}
\end{equation}

\subsection{Training Strategy}

Training is performed in three stages to progressively build from low-level visual representation learning to higher-level temporal reasoning and cross-modal alignment. First, the object detector from Section~\ref{sec:anatomicalregiondetection} is trained on Chest ImaGenome \cite{chest-imagenome-1.0.0}, a structured annotation resource derived from MIMIC-CXR, to localise the 29 predefined anatomical regions and extract region-level visual features. Second, the temporal modelling, clinical conditioning, and anatomical dropout modules are trained on paired current and prior studies from MIMIC-CXR \cite{MIMIC}, where longitudinal study pairs are identified using study identifiers, acquisition dates, and timestamps. During this stage, the detector is frozen to allow the model to learn longitudinal, region-aware representations grounded in clinical context without altering the anatomical detector.
In the final stage, the self-supervised multi-modal contrastive objective is introduced to align the selected temporal-clinical region representations with their corresponding region-level report descriptions. Unlike the previous stage, the full model is trained end-to-end, allowing the detector, temporal fusion module, clinical conditioning module, anatomical dropout module, projection heads, and contrastive text encoder to adapt jointly for cross-modal retrieval. To prioritise contrastive alignment, we assign a larger weight to the self-supervised multi-modal contrastive objective to prioritise learning a discriminative retrieval space, while retaining the object detection, region selection, and abnormality classification losses as auxiliary supervision. This final stage refines the shared embedding space while preserving the anatomical and longitudinal structure learned in the earlier stages.

\subsection{Inference}

At inference time, the model receives a current chest X-ray and its associated clinical indication, together with a prior study when available. The anatomical detector first localises the predefined anatomical regions and extracts region-level visual features. These features are then fused with the corresponding prior-region features through the temporal modelling module and conditioned on the clinical indication text, producing temporal-clinical anatomical representations.

The anatomical dropout module is used to determine which regions should contribute to the report. Specifically, the region selection head filters the full set of detected anatomical regions into a smaller set of report-relevant representations $\{\mathbf{b}_n\}_{n=1}^{N_{\text{sel}}}$. In parallel, the pathology head predicts whether each selected region is normal or abnormal. These pathology predictions are used to constrain retrieval.

For each selected region representation $\mathbf{b}_n$, the visual projection head maps the representation into the shared image-text embedding space. The resulting visual embedding is compared against a pre-computed index of (either normal or abnormal, depending on how the region was classified in the previous step) region-level sentence embeddings using cosine similarity. The highest-ranked sentence for each selected region is retrieved and the selected sentences are assembled to form the final report. This retrieval-based formulation enables efficient report generation efficient report generation without the need for very large autoregressive models and the challenge of hallucinations \cite{endo21,Yan2024}, while maintaining anatomical grounding, pathology-aware retrieval, and longitudinal consistency.

After retrieval, we apply a label-aware filtering step before forming the final report. Exact duplicate sentences are removed, and highly similar retrieved sentences are filtered to reduce redundancy. Structured labels associated with each retrieved sentence are then used to discard less informative sentences whose label sets are contained within a more specific retrieved sentence. Label polarity and finding groups are also used to reduce contradictory statements, particularly when positive and negative findings refer to the same or overlapping anatomical regions. When multiple retrieved sentences describe the same finding in the same or overlapping anatomical region, we prioritise sentences that contain explicit temporal cues, such as new, increased, improved, resolved, stable, or unchanged, over sentences that only describe the finding without longitudinal context. The remaining sentences are then assembled to form the final report.

\section{Experiments}
\noindent  {\bf Datasets --}
In this work we use MIMIC-CXR \cite{MIMIC} and its derivative, Chest ImaGenome \cite{chest-imagenome-1.0.0}. MIMIC-CXR \cite{MIMIC} is currently the largest publicly available chest x-ray dataset, containing 377,110 radiographic images from 227,835 radiology studies conducted on 65,379 individuals presenting to the Beth Israel Deaconess Medical Center Emergency Department. Each study has an associated radiology report which contains the following sections: indication, technique, comparison, findings and impression. 

The Chest ImaGenome dataset \cite{chest-imagenome-1.0.0} contains scene graphs for each frontal image within the MIMIC-CXR dataset. Each scene graph contains bounding box coordinates for 29 anatomical regions in the chest, as well as the sentences describing the findings and indications of each region if they exist in the corresponding radiology report. Clinical indications are extracted using the MIMIC-CXR \cite{MIMIC} section parsing tool, and prior studies are identified based on the StudyID, StudyDate, and StudyTime metadata fields. We adopt the official MIMIC-CXR split to preserve patient-level separation across training, validation, and testing, and to enable direct comparison with prior work.

\vspace*{2mm}

\noindent {\bf Implementation Details --}
The model is implemented in PyTorch, using pretrained models from the Transformers library. Albumentations is used for image preprocessing and augmentation. Following Tanida et al~\cite{tanida2023}, images are resized to $512 \times 512$ while preserving the orginal aspect ratio through padding where required. During training, we apply image augmentations including colour jitter, gaussian noise, and affine transformations.

Training is performed for 20 epochs with a batch size of 64 using the AdamW optimiser, with an initial learning rate of $1\times10^{-4}$ and weight decay of $1\times10^{-2}$. Learning rate updates are governed by a ReduceLROnPlateau scheduler, which applies multiplicative decay when the validation loss plateaus, down to a minimum learning rate of $1\times10^{-6}$ across all training phases. Training was performed on a single GH200 with 96Gb of vRAM. 

{For evaluation, we report retrieval performance using Recall@$k$, report-level lexical similarity using BLEU and ROUGE, and clinical correctness using CheXbert, RadCLiQ, and RadGraph F1. Recall@$k$ measures whether the relevant report or sentence appears among the top-ranked retrieved candidates, while BLEU and ROUGE assess surface-level overlap between generated and reference reports. Since lexical overlap does not necessarily reflect clinical correctness, we also use clinical metrics. CheXbert measures agreement between generated and reference reports across clinically important findings, such as pneumonia, cardiomegaly, edema, and pleural effusion, providing a label-based estimate of semantic correctness even when wording differs. RadGraph F1 evaluates agreement at the level of radiology entities and relations, capturing whether clinically meaningful observations and their relationships are preserved. RadCLiQ provides an aggregate estimate of clinical report error, where lower scores indicate greater clinical similarity to the reference report. Lexical and clinical metrics are computed using the RadEval framework~\cite{xu-etal-2025-radeval}. Together, these metrics capture complementary aspects of performance: retrieval accuracy, lexical similarity, clinical finding agreement, entity-relation correctness, and estimated clinical error. We repeat each experiment three times with different random seeds and report the mean.} 

\subsection{Results}
\begin{table*}[t]
\footnotesize
\centering
\resizebox{\textwidth}{!}{
\begin{tabular}{lccccccc} \hline
\textbf{Method} 
& \textbf{Publication} 
& \multicolumn{1}{c}{\textbf{Test Split}} 
& \multicolumn{5}{c}{\textbf{Recall@k}} \\[0.5ex]
\cline{4-8}
& & \multicolumn{1}{c}{\textbf{Sample Size}}
& \textbf{k=1} & \textbf{k=5} & \textbf{k=10} & \textbf{k=50} & \textbf{k=100} \\
\hline
 Ou et al. \cite{ou2025} & Multimedia Systems 2025 & 500 & 0.230 & 0.490 & 0.669 & - &- \\
 \framework & & 500 & \textbf{0.282} & \textbf{0.589} & \textbf{0.823} & - &- \\\hline
 BioViL-T$^\dagger$  \cite{Bannur2023} & CVPR 2023& MIMIC& 0.028& 0.111& 0.176& \underline{0.428}&\underline{0.582}\\ 
RadIR \cite{Zhang2025RadIR}& MICCAI 2025 & MIMIC & - & 0.043 & 0.069 & 0.181 & 0.213 \\
CLIP-XRad \cite{CLIP-XRad} & SYNASC 2025 & MIMIC & {0.038} & {0.162} & {0.259} & - & - \\
\framework &  & MIMIC & \textbf{0.197} & \textbf{0.494} & \textbf{0.616} & \textbf{0.818} & \textbf{0.869} \\ \hline
\end{tabular}
}
\vspace{0.5cm}
\caption{Performance against image-to-report retrieval methods on MIMIC-CXR using Recall@k metrics. The Test Split Sample Size column specifies the evaluation subset size used for each comparison; \framework rows are matched to the corresponding referenced setting where applicable. The best-performing approach within each setting is shown in bold. $^\dagger$ indicates results obtained by re-evaluating the released BioViL-T~\cite{Bannur2023} model with our evaluation pipeline.}
\label{tbl:recallk}
\end{table*}

Table~\ref{tbl:recallk} compares \framework against existing image-to-report retrieval methods on MIMIC-CXR using Recall@k. Prior methods report results under different evaluation settings, either using a fixed-size sampled subset or the official MIMIC-CXR test split. To enable fairer comparison, we reproduce each setting as closely as possible and report the corresponding evaluation setting in the table.

In the 500-sample setting used by Ou et al.~\cite{ou2025}, \framework improves Recall@1 by 0.052, Recall@5 by 0.099, and Recall@10 by 0.154. With the official MIMIC-CXR split used by Rad-IR~\cite{Zhang2025RadIR} and CLIP-XRad~\cite{CLIP-XRad}, \framework also achieves stronger performance across all reported retrieval depths, with particularly large gains at Recall@50 and Recall@100 compared with Rad-IR. These results suggest that the proposed anatomically grounded and temporally informed retrieval strategy improves both top-ranked retrival accuracy and broader candidate recall.

\begin{table*}[ht!]
\centering
\footnotesize
\setlength{\tabcolsep}{4pt}
\begin{tabular}{llcccccc}
\hline
\textbf{Model} & \textbf{Publication} &  \textbf{BLEU-1}&\textbf{BLEU-2}  &\textbf{BLEU-3}& \textbf{BLEU-4}  &\textbf{BLEU-Avg}& \textbf{ROUGE-L} \\[0.5ex]
\hline
CxR-RepaiR \cite{endo21} & ML4H 2021 &  -&0.055  &-& 0.021  &-& 0.143 \\
BioViL \cite{Boecking2022} & ECCV 2022 &  -&-  &-& 0.037  & - & 0.200 \\
BioViL-T \cite{Bannur2023} & CVPR 2023 & 0.357$\dagger$& 0.216$\dagger$  & 0.149$\dagger$ & 0.045  &0.192$\dagger$& 0.205\\
X-REM \cite{Jeong2023MIDL} & MIDL 2023 &  -& 0.084&-& -  &-& - \\
Teaser  \cite{Teaser} & TMI 2024 &  \underline{0.423} & 0.257  & 0.166 & 0.113  & 0.240
& \textbf{0.287}\\
AHIVE \cite{Yan2024} & CVPR 2024 & -& -  &-& -&0.131& 0.163 \\
DuCo-Net \cite{UrRahman2025} & IEEE Access 2025 & 0.420 & \underline{0.270} & \underline{0.170} & \underline{0.120} & \underline{0.245} & \underline{0.220} \\
\framework & & \textbf{0.437} & \textbf{0.281} & \textbf{0.172} & \textbf{0.128}& \textbf{0.249} & 0.218 \\
\hline
\end{tabular}%
\vspace{0.3cm}
\caption{Performance of image-to-report retrieval methods on MIMIC-CXR using standard natural language generation metrics. Best approach for each metric is highlighted in bold and the second best is underlined. $^\dagger$ denotes results reproduced with the released BioViL-T~\cite{Bannur2023} model using our evaluation pipeline.}
\label{tbl:nlp_results}
\end{table*}

Table~\ref{tbl:nlp_results} reports standard natural language generation metrics for image-to-report retrieval methods on the official MIMIC-CXR test split. \framework achieves the strongest performance across all BLEU metrics, obtaining BLEU-1, BLEU-2, BLEU-3, and BLEU-4 scores of 0.437, 0.281, 0.172, and 0.128, respectively. Compared with the strongest prior result for each metric, \framework improves BLEU-1 by 0.014 over Teaser and improves BLEU-2, BLEU-3, BLEU-4, and BLEU-Avg by 0.011, 0.002, 0.008, and 0.004 over DuCo-Net. Although these gains are modest, they indicate consistently stronger lexical and phrase-level overlap with reference reports across unigram to 4-gram matching.

In contrast, \framework obtains a ROUGE-L score of 0.218, which is lower than Teaser~\cite{Teaser}. This suggests that, while \framework improves phrase-level precision, as reflected by BLEU, it may not capture the exact wording of reference-report content measured by ROUGE. Lexical overlap metrics do however, only provide a partial view of radiology report quality \cite{Sloan2024}, as clinically correct reports may use sentences of semantic equivalence despite different wording. We therefore complement these lexical results with clinical efficacy metrics reported in Table~\ref{tbl:clinical_results}. 
which demonstrates that \framework achieves substantially stronger clinical performance than prior retrieval-based methods across all reported clinical metrics. On CHEXBERT, \framework obtains the best score of 0.713, improving over AHIVE by 0.028. On the other clinical metrics, X-REM~\cite{Jeong2023MIDL} is the strongest prior baseline where \framework improves RadGraph F1 by 0.028 and reduces RadCliQ from 3.781 to 0.569. Since RadCliQ estimates clinical report error, lower values indicate better performance \cite{Jeong2023MIDL}. {Together, these results demonstrate the ability of \framework to retrieve reports that are not only textually similar to references but also more clinically faithful than those retrieved by prior methods.}

\begin{table*}[t!]
\centering
\footnotesize
\setlength{\tabcolsep}{4pt}
\begin{tabular}{llccc}
\hline
\textbf{Model} & \textbf{Publication} & \textbf{CHEXBERT} & \textbf{RadCLiQ $\downarrow$} & \textbf{RadGraph F1} \\[0.5ex]
\hline
CxR-RepaiR \cite{endo21} & ML4H 2021 & 0.274& 4.121 & 0.09\\
BioViL \cite{Boecking2022} & ECCV 2022 & 0.283 & - & - \\
BioViL-T \cite{Bannur2023} & CVPR 2023 & 0.290 & 0.645$\dagger$ & 0.138$\dagger$ \\
X-REM \cite{Jeong2023MIDL} & MIDL 2023 & 0.381& \underline{3.781} & \underline{0.181}\\
 Teaser \cite{Teaser}& TMI 2024& 0.526& -&-\\
 AHIVE \cite{Yan2024}& CVPR 2024& \underline{0.685} & - &-\\
\framework & - & \textbf{0.713}& \textbf{0.569}& \textbf{0.209}\\
\hline
\end{tabular}%
\vspace{0.3cm}
\caption{Performance of image-to-report retrieval methods on MIMIC-CXR using clinical correctness and radiology-specific evaluation metrics. Metrics denoted with $\downarrow$ indicate that lower scores are better. Best method's result for each metric is in bold while second best is underlined. $^\dagger$ means results are reproduced using the released BioViL-T \cite{Bannur2023} model and our evaluation pipeline.}
\label{tbl:clinical_results}
\end{table*}


\subsection{Ablations}
To evaluate the contribution of the proposed architectural components, a series of ablation studies were conducted examining both the anatomical region localisation framework and the multimodal temporal fusion strategies used for region-level classification and retrieval.
\begin{table*}[b!]
\footnotesize
\centering
\begin{tabular}{lccccc} \hline
\textbf{Model} & \textbf{mAP@0.5} & \textbf{Mean IoU} & \textbf{Precision@0.5} & \textbf{Recall@0.5} & \textbf{F1@0.5} \\[0.5ex]
\hline
Faster R-CNN \cite{FasterRCNN} & 0.87 & 0.89 & 0.90 & 0.86 & 0.88 \\
Faster R-CNN with RadDINO & \textbf{0.93} & \textbf{0.94} & \textbf{0.95} & 0.92 & \textbf{0.93} \\
DETR \cite{detr} & 0.84 & 0.87 & 0.88 & 0.83 & 0.85 \\
YOLOv10 \cite{yolov10} & \underline{0.91} & \underline{0.92} & \underline{0.92} & \textbf{0.93} & \underline{0.92} \\
\hline
\end{tabular}
\vspace{0.5cm}
\caption{Quantitative evaluation of anatomical region localization using mAP@0.5, Mean IoU, Precision@0.5, Recall@0.5, and F1@0.5. The best-performing method is shown in bold and the second best is underlined.}
\label{tbl:object_detection_results}
\end{table*}

The quantitative results in Table~\ref{tbl:object_detection_results} demonstrate strong anatomical region localisation performance across all evaluated object detection architectures. To identify the most effective detector for anatomical region localisation in chest radiographs, several widely used object detection architectures were evaluated, including Faster R-CNN \cite{FasterRCNN}, DETR \cite{detr}, and YOLOv10 \cite{yolov10} approaches. Among the evaluated methods, the proposed RadDINO-based transformer detector achieved the best overall performance, obtaining a mAP@0.5 of 0.93 and a Mean IoU of 0.94, while also achieving the highest Precision@0.5 and F1@0.5 scores. These findings suggest that radiology-specific transformer representations substantially improve localisation accuracy and overall detection robustness for anatomical region detection.

Among the baseline approaches, YOLOv10 achieved the second-best overall performance, outperforming the standard Faster R-CNN and DETR architectures across most evaluation metrics. YOLOv10 reached the highest Recall@0.5 score of 0.93, indicating improved sensitivity for anatomical region detection and fewer missed structures. Faster R-CNN demonstrated stable and balanced performance, while DETR produced comparatively lower localisation accuracy and recall. Overall, the results indicate that transformer-based radiology representations provide strong advantages for anatomical region localisation, particularly when combined with domain-specific pretraining such as RadDINO.

\begin{table*}[t]
\scriptsize
\centering
\resizebox{0.94\textwidth}{!}{
\begin{tabular}{lcccc} \hline
\textbf{Method} & \textbf{Selection AP} & \textbf{Selection ROC-AUC} & \textbf{Abnormality AP} & \textbf{Abnormality ROC-AUC} \\[0.5ex]
\hline
Additive Conditioning & 0.761 & 0.919 & 0.680 & 0.906 \\
FiLM Conditioning & \underline{0.908} & \underline{0.952} & \textbf{0.869} & \textbf{0.975} \\
Region-to-Text Cross-Attention (MHA) & \textbf{0.921} & \textbf{0.959} & \underline{0.866} & \underline{0.969} \\
\hline
\end{tabular}
}
\vspace{0.5cm}
\caption{Ablation study evaluating different multimodal temporal fusion strategies for region selection and abnormality classification. Best-performing results are shown in bold and second-best results are underlined.}
\label{tbl:temporal_ablation_results}
\end{table*}

The ablation results in Table~\ref{tbl:temporal_ablation_results} demonstrate the impact of multimodal fusion strategies for temporal anatomical region classification. To investigate the most effective mechanism for integrating clinical indication text with temporally fused anatomical region representations, several fusion approaches were explored. The first approach employed simple additive conditioning, where a global RadBERT embedding was directly added to the temporally fused visual features. A second approach utilised FiLM-based conditioning \cite{perez2018_film}, where the textual representation was used to adaptively modulate the visual embeddings. Finally, a region-to-text cross-attention strategy was investigated, enabling each anatomical region representation to attend directly to individual textual token embeddings generated by RadBERT.

Among the evaluated approaches, additive conditioning produced the weakest overall performance, while FiLM conditioning improved both region selection and abnormality classification performance. The proposed region-to-text cross-attention formulation achieved the strongest overall region selection performance, obtaining the highest Selection AP and Selection ROC-AUC scores. Although FiLM conditioning achieved marginally stronger abnormality classification performance, region selection performance was prioritised during model selection, as this component determines whether clinically relevant anatomical regions are selected for downstream retrieval and report generation. Consequently, the cross-attention model was selected as the architecture due to its stronger semantic grounding and improved ability to consistently retrieve anatomically relevant region descriptions.

\begin{table*}[ht!]
\footnotesize
\centering
\begin{tabular}{lccccc}
\hline
\textbf{Method} & \multicolumn{5}{c}{\textbf{Recall@$k$}} \\
\cline{2-6}
& \textbf{$k=1$} & \textbf{$k=5$} & \textbf{$k=10$} & \textbf{$k=50$} & \textbf{$k=100$} \\
\hline
\framework~EM & 0.197 & 0.494 & 0.616 & 0.818 & 0.869  \\
\framework~SM & 0.746 & 0.877 & 0.905 & 0.941 & 0.950 \\
\hline
\end{tabular}
\vspace{0.5cm}
\caption{Ablation demonstrating the Recall@$k$ performance of \framework using exact-match (EM) and semantic-match (SM) retrieval on MIMIC-CXR.}
\label{tbl:abl_EMvsSM}
\end{table*}

Table~\ref{tbl:abl_EMvsSM} compares exact-match (EM) and semantic-match (SM) retrieval. EM requires retrieved region-level text to match the reference text exactly, making it a strict retrieval measure. In contrast, SM uses the structured Chest ImaGenome~\cite{chest-imagenome-1.0.0} labels associated with each region-level sentence. A candidate is counted as a semantic match when its structured Chest ImaGenome labels fully overlap with the ground-truth region label set, allowing clinically equivalent sentences to be treated as correct even when their wording differs.

The large gap between EM and SM, particularly at low $k$ values, demonstrates that many retrievals judged incorrect under exact textual matching are nevertheless clinically meaningful. This supports the difference observed between lexical and clinical metrics, as surface-level overlap may penalise clinically valid alternative phrasings, while label-based semantic matching better reflects preservation of the underlying clinical findings. This suggests that the improved clinical efficacy of \framework arises from retrieving region-level descriptions that preserve the same structured clinical labels as the reference, despite different wording.

\section{Conclusions}
In this work, we introduce \framework, a spatio-temporal and clinically conditioned retrieval framework for automated radiology report generation. Unlike conditional retrieval-based approaches that operate primarily at the global image-report level, \framework performs sentence retrieval at the anatomical-region level, allowing report content to be grounded in visual findings. This design more closely reflects the radiologists' working practice \cite{tanida2023, sloan2025camchex}, where findings are described with respect to anatomical location, temporal change, and the clinical question motivating the examination. Experiments on MIMIC-CXR demonstrate that \framework outperforms existing retrieval-based baselines across retrieval, lexical and clinical evaluation metrics, with ablation studies further supporting the efficacy of anatomically grounded, temporally and clinically informed retrieval for radiology report generation.

Despite these improvements, several limitations remain. First, the retrieval-based formulation constrains grounding to sentences present in the retrieval corpus, which reduces hallucination risk but may limit the ability to express rare, complex, or previously unseen findings. Secondly, \framework currently operates on frontal chest X-rays and does not explicitly incorporate lateral views or multi-view reasoning, which may limit its ability to capture findings that are more clearly visible from complementary projections. 

Future work will investigate more flexible retrieval and report composition strategies, incorporating multi-view inputs and investigating the effectiveness of extending temporal modelling beyond a single prior study. In addition to these extensions, one promising research direction is to develop \framework as part of a retrieval-augmented reporting workflow, where the retrieved anatomically grounded sentences provide evidence for downsteam report drafting, editing or verification. This could also be expanded towards an agentic solution, which retrieves candidate findings, checks for missing anatomical regions, pathologies or contradictory statements and refines the report structure. 

Interactivity and human-in-the-loop functionality represent a natural extension of \framework. Since retrieval is performed at the anatomical-region level, radiologists could guide report generation by selecting or refining bounding boxes, confirming clinically relevant regions, or choosing from retrieved candidate sentences. This would allow expert users to retain control over the final report while using \framework as an anatomically grounded retrieval aid.


\bibliography{bibliography}

\end{document}